\documentclass[10pt,twocolumn,letterpaper]{article}

\usepackage{cvpr}              

\usepackage[accsupp]{axessibility}  

\usepackage{graphicx}
\usepackage{amsmath}
\usepackage{amssymb}
\usepackage{booktabs}

\usepackage{comment}
\usepackage{float}  
\usepackage{multirow}
\usepackage{tabularx}
\usepackage{makecell}

\usepackage{verbatim}

\usepackage{amsfonts}
\usepackage{bm}

\usepackage{color}

\usepackage[pagebackref,breaklinks,colorlinks]{hyperref}

\usepackage[capitalize]{cleveref}
\crefname{section}{Sec.}{Secs.}
\Crefname{section}{Section}{Sections}
\Crefname{table}{Table}{Tables}
\crefname{table}{Tab.}{Tabs.}

\def\cvprPaperID{4952} 
\def\confName{CVPR}
\def\confYear{2022}

\begin{document}

\title{EMScore: Evaluating Video Captioning via \\ Coarse-Grained and Fine-Grained Embedding Matching}

\author{ Yaya Shi$^1$, Xu Yang$^2$, Haiyang Xu$^3$, Chunfeng Yuan$^4$\thanks{Corresponding author}\;, Bing Li$^4$, Weiming Hu$^{4,5,6}$, Zheng-Jun Zha$^1$\\
$^1$University of Science and Technology of China~~~
$^2$Southeast University~~~
$^3$Alibaba Group~~~ \\
$^4$NLPR, Institute of Automation, Chinese Academy of Sciences \\
$^5$School of Artificial Intelligence, University of Chinese Academy of Sciences \\
$^6$CAS Center for Excellence in Brain Science and Intelligence Technology \\
{\tt\small shiyaya@mail.ustc.edu.cn~~~101013120@seu.edu.cn~~~shuofeng.xhy@alibaba-inc.com} \\ 
{\tt\small \{cfyuan, bli, wmhu\}@nlpr.ia.ac.cn~~~zhazj@ustc.edu.cn}
}

\maketitle

\begin{abstract}
	Current metrics for video captioning are mostly based on the text-level comparison between reference and candidate captions. However, they have some insuperable drawbacks, \eg, they cannot handle videos without references, and they may result in biased evaluation due to the one-to-many nature of video-to-text and the neglect of visual relevance. From the human evaluator's viewpoint, a high-quality caption should be consistent with the provided video, but not necessarily be similar to the reference in literal or semantics. Inspired by human evaluation, we propose \textbf{EMScore} (Embedding Matching-based score), a novel reference-free metric for video captioning, which directly measures similarity between video and candidate captions. Benefiting from the recent development of large-scale pre-training models, we exploit a well pre-trained vision-language model to extract visual and linguistic embeddings for computing EMScore. Specifically, EMScore combines matching scores of both coarse-grained (video and caption) and fine-grained (frames and words) levels, which takes the overall understanding and detailed characteristics of the video into account. Furthermore, considering the potential information gain, EMScore can be flexibly extended to the conditions where human-labeled references are available. Last but not least, we collect VATEX-EVAL and ActivityNet-FOIl datasets to systematically evaluate the existing metrics. VATEX-EVAL experiments demonstrate that EMScore has higher human correlation and lower reference dependency. ActivityNet-FOIL experiment verifies that EMScore can effectively identify ``hallucinating'' captions. Code and datasets are available at~\url{https://github.com/shiyaya/emscore}.
\end{abstract}

\begin{figure}[t]
	\centering
	\setlength{\abovecaptionskip}{1mm}
	\includegraphics[width=0.8\linewidth]{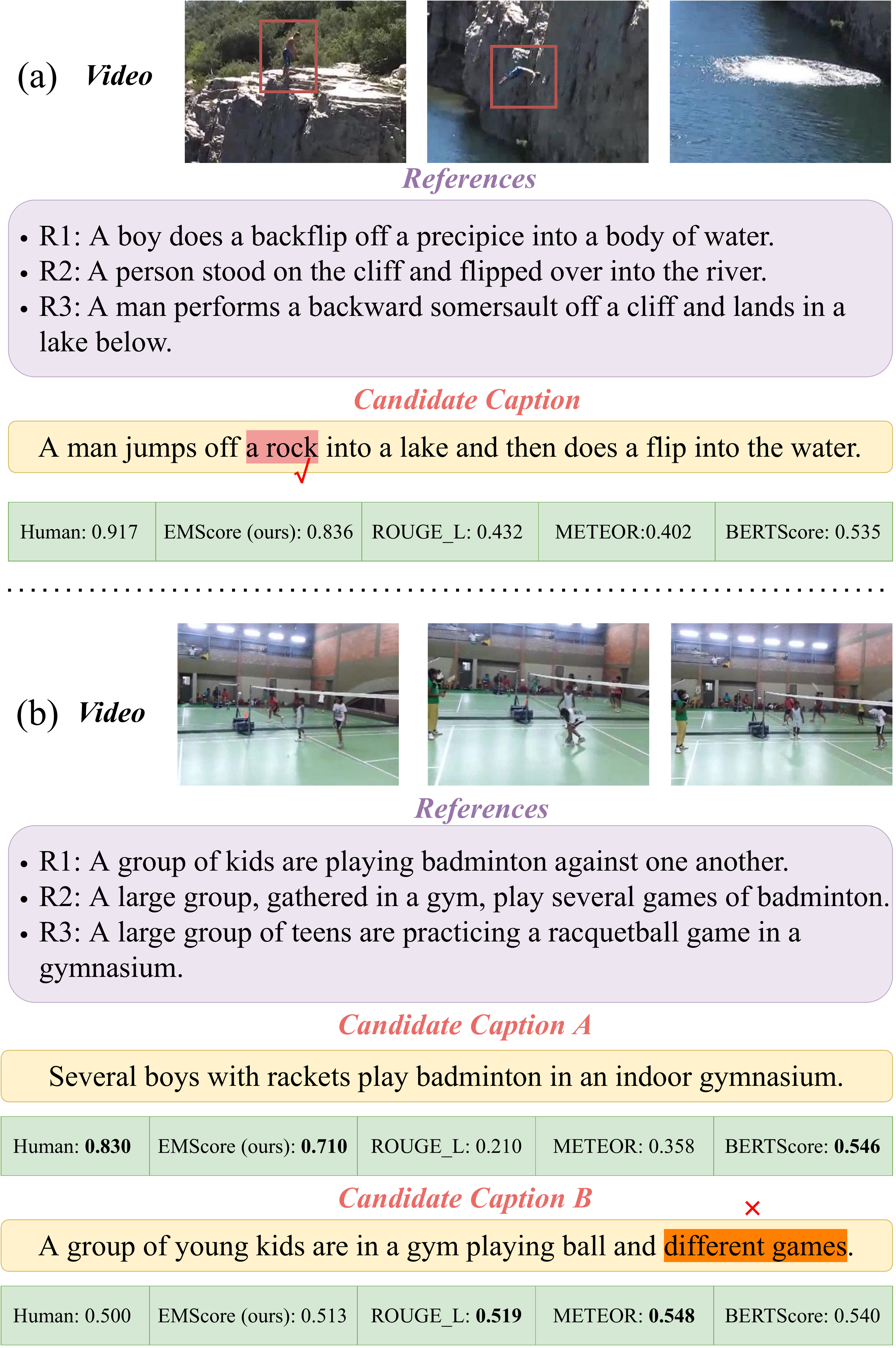}
	\caption{
		Two examples of caption evaluation. 
		All the metric scores are scaled to [0, 1], including human scores. 
		For example (a), reference-based metrics over-penalize for this correct candidate caption due to ``a rock'' is not contained in the references. Our reference-free metric \text{EMScore} gives a reasonable high score with the help of using video as ground truth.
		For example (b), some reference-based metrics (e.g., ROUGE\_L and METEOR) under-penalize the hallucination (e.g., ``different games'') which is not related to the video, and give an unreasonable higher score for ``hallucinating'' caption B than correct caption A.
	}
	\label{fig:introductionexample}
	\vspace{-5mm}
\end{figure}


\section{Introduction}
\label{sec:intro}
Video Captioning~\cite{DBLP:journals/tcsv/DengLZWZH22} aims to generate a text describing the visual content of a given video. 
Driven by the neural encoder-decoder paradigm, research in video captioning has made significant progress~\cite{DBLP:conf/iccv/VenugopalanRDMD15, DBLP:conf/cvpr/ZhangSY0WHZ20}. 
To make further advances in video captioning, it is essential to accurately evaluate generated captions. The most ideal metric is human evaluation while carrying human judgments is time-consuming and labor-intensive. Thus, various automatic metrics are applied for video caption evaluation.

However, most of the widely applied video caption metrics like BLEU~\cite{DBLP:conf/acl/PapineniRWZ02}, ROUGE~\cite{lin-2004-rouge}, CIDEr~\cite{7299087}, and BERTScore~\cite{DBLP:conf/iclr/ZhangKWWA20} come from the other tasks, such as machine translation, text summarization and image captioning, which may neglect the special characteristic of video captioning and then limit the development of video captioning. Furthermore, these automatic metrics require human-labeled references --- and thus they are called reference-based metrics --- and such requirements cause three intrinsic drawbacks: 
(1) They can not be used when provided videos have no human-labeled references, which is not uncommon in this age that millions of reference-free videos are produced online every day.
(2) They may over-penalize the correct captions since references hardly describe all details of videos due to the one-to-many nature~\cite{DBLP:conf/acl/YiDH20} of captioning task, especially when the number of references is limited. Fig.\ref{fig:introductionexample} (a) shows one such example where a candidate caption correctly describes the ``a rock'' while reference-based metrics punish this word since references do not contain it.
(3) As pointed by~\cite{rohrbach-etal-2018-object}, these reference-based metrics may under-penalize the captions with ``hallucinating'' descriptions since these metrics only measure similarity to references, and the visual relevance cannot be fully captured. For example, as shown in Fig.\ref{fig:introductionexample} (b), due to the word ``games'' appearing in the references, some reference-metrics return higher scores for caption B than caption A, even though ``different games'' is a ``hallucinating'' phrase which is not related to the video.

These drawbacks inspire us to develop a reference-free metric. 
From the human evaluator's viewpoint, if a caption is \textit{consistent} with the source video, \emph{i.e.}, the visual contents in the video are comprehensively and accurately described by the caption, this caption is a high-quality one, and not necessarily be similar to the reference in literal or semantics. A promising evaluation metric should imitate the human evaluation process, and introduce video content into the evaluation. Nowadays, due to the boom of the large-scale vision-language pre-training models~\cite{DBLP:conf/icml/RadfordKHRGASAM21, DBLP:conf/emnlp/LiCCGYL20, DBLP:conf/cvpr/MiechASLSZ20}, the gaps between the visual and linguistic embeddings have been further narrowed, enabling us to judge whether a caption is consistent with a video. 

Motivated by these research progresses, we propose a \textit{reference-free} metric \text{\textbf{EMScore}} (Embedding Matching-based score) for evaluating video captions, which exploits a pre-trained large-scale vision-language model to extract visual and linguistic embeddings. Specifically, to obtain a comprehensive comparison between the video and caption, EMScore averages the matching scores of both coarse-grained (video and caption) and fine-grained (frames and words) levels. 
For the coarse-grained one, we compute the similarity between the global embeddings of the video and the candidate caption, which take the overall understanding of the video into account and evaluate candidates from a global perspective.
For the fine-grained embedding matching, we compute the sum of cosine similarities between the frame and word embeddings, which takes the detailed characteristic of the video (visual elements change over time) into account. Also, it provides more interpretability for EMScore.
Furthermore, considering the potential information gain, such as syntactic structure in references, and doing embedding matching in the same language domain is easier than cross-modal domains, we extend EMScore to the conditions where human-labeled references are available and name the extended metric \text{EMScore\_ref}.

Currently, there is no available video caption quality dataset that can be used to evaluate metrics. To facilitate the development of video captioning evaluation metrics, we are the first to collect a video caption quality dataset VATEX-EVAL which contains 54,000 human ratings for video-caption pairs.  
Experiments on VATEX-EVAL show the following advantages of our EMScore by introducing the video in evaluating. 
First, EMScore has a higher human correlation compared with some popular automatic metrics like BLEU, ROUGE, or CIDEr.
Second, EMScore has low reference dependency, \emph{e.g.}, EMScore's 0-reference Kendall's correlation with humans is similar to BLEU\_1's 4-reference correlation or EMScore\_ref's 1-reference is similar to CIDEr's 9-reference correlations. Therefore, EMScore can significantly reduce the cost of manually annotating references.
Third, EMScore is more robust to quality drift that it achieves higher correlations compared with the other automatic metrics when evaluating captions of different qualities.
Furthermore, we collect another dataset ActivityNet-FOIL which contains ``hallucinating'' captions to verify the sensitivity of EMScore. Experiment results show that EMScore is more effective to identify ``hallucinating'' captions than the other metrics.

Our contributions are summarized as follows: 
\vspace{-3mm}
\begin{itemize}
	\setlength{\itemsep}{0pt}
	\setlength{\parsep}{0pt}
	\setlength{\parskip}{0pt}
	\item  We propose a reference-free video captioning metric EMScore that directly measures consistency with video contents in both coarse-grained and fine-grained levels, and extend it to reference-available condition.
	\item We collect two datasets VATEX-EVAL and ActivityNet-FOIL for researchers to study the metrics' correlation with human judgments and sensitivity in the ``hallucinating'' case, respectively. 
	\item Exhaustive experimental results verify that EMScore has a higher human correlation and is able to effectively identify the ``hallucinating'' captions.
\end{itemize}

\section{Related work}
\vspace{-1mm}
\subsection{Caption Evaluation}
\vspace{-1mm}



\noindent\textbf{Rule-Based Evaluation} 
The most widely used caption metrics are based on n-gram matching --- BLEU~\cite{DBLP:conf/acl/PapineniRWZ02}, ROUGE~\cite{lin-2004-rouge} and METEOR~\cite{banerjee-lavie-2005-meteor}. Especially, CIDEr~\cite{7299087} weights each n-gram by tf-idf.
However, they are sensitive to lexical variation and hard to capture semantics of a caption, so they correlate poorly with human judgments~\cite{DBLP:conf/iclr/ZhangKWWA20}.


\noindent\textbf{Embedding-Based Evaluation} 
Embedding-based metrics which use pre-trained models to extract embeddings and perform semantic matching in the embedding space, have been proven to correlate better with human judgments.
BERTScore~\cite{DBLP:conf/iclr/ZhangKWWA20} uses contextual word embeddings generated by BERT, and measures the semantic similarity of two texts by computing token-level cosine similarity. 
BERTScore can be regarded as a special case of ours, it only uses references for evaluation and performs single fine-grained embedding matching.
Among these embedding metrics, some works try to take into account the vision information.
Tiger~\cite{DBLP:conf/emnlp/JiangHZWZGDG19} uses a trained image-text matching SCAN model~\cite{DBLP:conf/eccv/LeeCHHH18} to compare the ground outputs between candidate caption and reference. 
ViLBERTScore~\cite{lee-etal-2020-vilbertscore} uses a pre-trained ViLBERT model~\cite{DBLP:conf/nips/LuBPL19} to compare the visually-grounded text representation between candidate caption and reference.
In these two evaluation metrics, the image is used as a visual ground in the evaluation rather than as ground truth, and they are still reference-based metrics. 
CLIPScore~\cite{DBLP:conf/emnlp/HesselHFBC21} and FAIEr~\cite{DBLP:conf/cvpr/WangYWWC21} are recently proposed reference-free evaluation metrics. CLIPScore~\cite{DBLP:conf/emnlp/HesselHFBC21} uses the pre-trained image-language model CLIP~\cite{DBLP:conf/icml/RadfordKHRGASAM21} to obtain image and text embeddings, and compute the cosine similarity. But they only consider coarse-grained matching and ignore fine-grained ones, so that CLIPScore lacks interpretability and ignores that a more precise score comes from fine-grained matching.
FAIEr~\cite{DBLP:conf/cvpr/WangYWWC21} introduces the scene graph to evaluate the fidelity and adequacy of the image captions. 
The above metrics are all proposed for image captioning.
In this paper, we propose an evaluation metric specifically for video captioning by introducing video content. We consider not only coarse-grained embedding matching between video and text but also the fine-grained embedding matching between frames and words to take into account the characteristic of the visual elements of the video over time.

\vspace{-1mm}
\subsection{Pre-trained Vision-Language Models}
\vspace{-1mm}

Inspired by the success of the large-scale pre-training in NLP~\cite{DBLP:conf/naacl/DevlinCLT19, radford2019language}, large-scale pre-training models~\cite{DBLP:conf/emnlp/TanB19, DBLP:conf/nips/LuBPL19, DBLP:conf/emnlp/LiCCGYL20, DBLP:conf/cvpr/MiechASLSZ20, xu-etal-2021-e2e} also become the research hotspot in the vision-language community. Generally, these models are pre-trained by pretext tasks on large-scale datasets, such as Conceptual captions~\cite{DBLP:conf/acl/SoricutDSG18} and HowTo100M~\cite{DBLP:conf/iccv/MiechZATLS19} . During the pre-training, the models learn to narrow the gaps between the vision and language embeddings, which enables them to generalize well to various down-stream tasks like VQA~\cite{DBLP:conf/iccv/AntolALMBZP15}, Visual Grounding~\cite{DBLP:conf/iccv/LiuZZW19}, Image/Video-Text retrieval and Image/Video Captioning~\cite{DBLP:conf/mm/LiuZZZW18, DBLP:journals/pami/ZhaLZZW22, DBLP:conf/ijcai/TanLWZ20}. Motivated by the narrowed embedding gaps, we exploit one large-scale pre-trained model: CLIP~\cite{DBLP:conf/icml/RadfordKHRGASAM21}, which is pre-trained via contrastive learning on 400 million image-text pairs, to design a video caption metric. CLIP-straight~\cite{DBLP:journals/corr/abs-2102-12443} shows that straight forward applying CLIP to video-text retrieval can achieve excellent zero-shot performance, which proves that the gaps between the extracted video and text embeddings are reduced.
Therefore, by CLIP, measuring the consistency between the video content and the candidate caption is transformed to computing the cosine similarity between the extracted video and caption embeddings.

\begin{figure*}[t]
	\centering
	\setlength{\abovecaptionskip}{1mm}
	\includegraphics[width=0.65\linewidth]{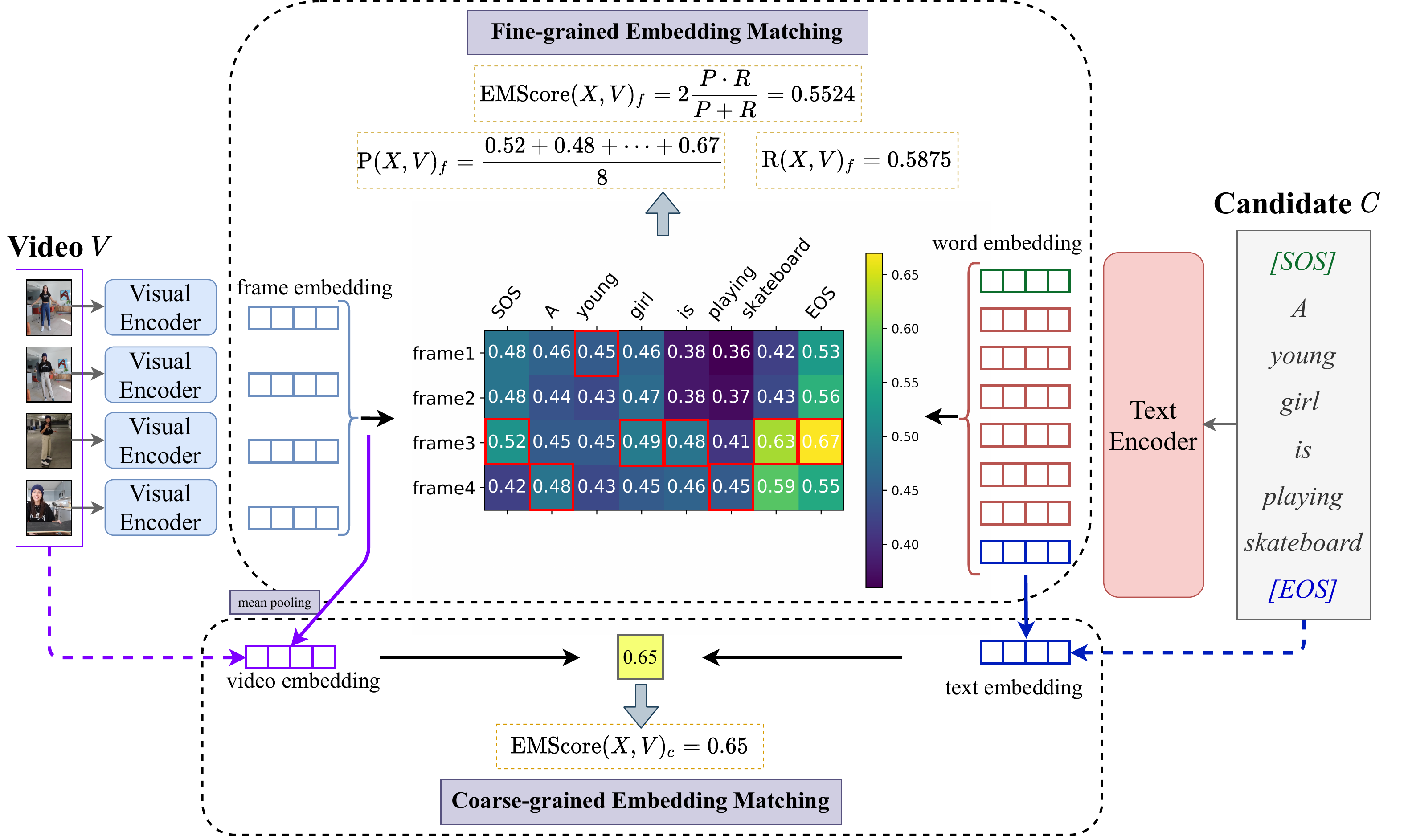}
	\caption{Illustration of the computation of the EMScore which uses video as ground truth. Given the video $V$ and candidate caption $C$, 
		we extract global representations of video and caption for coarse-grained vector matching $\text{EMScore(X,V)}_{c}$, and local representations of frames and words for fine-grained greedy matching $\text{EMScore(X,V)}_{f}$. We highlight the precision greedy matching in red, and for simplicity, we give the calculation without idf weighting. The overall EMScore is the average score of $\text{EMScore(X,V)}_{c}$ and $\text{EMScore(X,V)}_{f}$.}
	\label{fig: Model}
	\vspace{-5mm}
\end{figure*}

\vspace{-1mm}
\section{EMScore}
\vspace{-1mm}

Fig.\ref{fig: Model} shows the pipeline of EMScore, which computes the embedding similarity of the generated captions and the source video to achieve reference-free caption evaluation. 

\vspace{-1mm}
\subsection{Embedding Extraction}
\vspace{-1mm}

\label{subsec:embedding_extraction}
We use CLIP~\cite{DBLP:conf/icml/RadfordKHRGASAM21} to extract video and text embeddings at both fine-grained and coarse-grained levels. Specifically, the visual encoder $E_v$ (ViT-B/32)~\cite{DBLP:conf/iclr/DosovitskiyB0WZ21} extracts the embeddings of individual frame and total video. The language encoder $E_t$ (Transformer)~\cite{radford2019language} extracts the embeddings of each token and whole sentence.

\noindent\textbf{Frame and Video Representation}
Given a video $V=\{v_i\}_{i=1}^{|V|}$ ($|V|$ is the number of frames), each fine-grained frame embedding $f_{v_{i}}$ is obtained as follows:
\vspace{-2mm}
\begin{equation}
	\begin{footnotesize}
		\mathbf{f}_{v_i}=\operatorname{Norm}\left(E_{v}(v_i)\right), \mathbf{f}_{v} \in \mathbb{R}^{d},
	\end{footnotesize}
	\vspace{-3mm}
\end{equation}
where Norm$(\cdot)$ is a L2 normalization function.

The coarse-grained video embedding $\mathbf{f}_{V}$ is the normalization of the mean-pooling of all the frame embeddings:
\vspace{-2mm}
\begin{equation}
	\begin{footnotesize}
		\mathbf{f}_{V}=\operatorname{Norm}\left(\frac{1}{|V|} \cdot \sum_{i=1}^{|V|} \mathbf{f}_{v_{i}} \right), \mathbf{f}_{V} \in \mathbb{R}^{d}.
	\end{footnotesize}
	\vspace{-2.5mm}
\end{equation}

\noindent\textbf{Word and Text Representation}
Given a caption, we first use the default tokenizer of CLIP to obtain word tokens and then add two special tokens [SOS] and [EOS] to construct a new token sequence  $X=\left\{x_{j}\right\}_{j=1}^{|X|}$ ($|X|$ is the number of tokens). 
The contextual token embeddings are:
\vspace{-2mm}
\begin{equation}
	\label{equation: text representation}
	\begin{footnotesize}
		\begin{aligned}
			& \{\mathbf{f}_{sos}, \mathbf{f}_{x_{1}}, \cdots, \mathbf{f}_{x_{|X|-2}},\mathbf{f}_{eos}\} =\operatorname{Norm}\left( W \cdot LN(E_{t}(X)) \right),  \\
			& \mathbf{f}_{x} \in \mathbb{R}^{d} ,
		\end{aligned} 
	\end{footnotesize}
	\vspace{-3mm}
\end{equation}
where LN is Layer Normalization, $W \in \mathbb{R}^{h \times d}$ are fixed parameters from CLIP, and $h$ is the hidden size of text encoder. 
All these $|X|$ token embeddings are used for fine-grained embedding matching and the last $\mathbf{f}_{eos}$ is treated as the global embedding $\mathbf{f}_{X}$ for coarse-grained embedding matching.

\vspace{-1mm}
\subsection{Embedding Matching}
\vspace{-1mm}

\label{subsec:embedding_matching}
\noindent\textbf{Coarse-grained Embedding Matching}
Given the source video $V$ and the generated caption $X$, the coarse-grained embedding matching $\text{EMScore}(X, V)_{c}$ is:
\vspace{-2mm}
\begin{equation}
	\begin{footnotesize}
		\text{EMScore}(X, V)_{c} = \mathbf{f}_{X}^{\top} \mathbf{f}_{V}, \footnote{Since all the embeddings are L2 normalized, the cosine similarity is reduced to the inner product.}
	\end{footnotesize}
	\vspace{-3mm}
\end{equation}
where $\mathbf{f}_{V}$ and $\mathbf{f}_{X}$ are embeddings of the video and caption, respectively. Such process is shown in lower part of Fig.~\ref{fig: Model}.

\noindent\textbf{Fine-grained Embedding Matching}
For videos, since visual elements in the frame change over time, only performing the coarse-grained embedding matching may lose detailed information, which inspires us to design a fine-grained embedding matching to achieve frame-token alignment.
The upper part of Fig.\ref{fig: Model} shows the applied fine-grained matching. Given the video frame embedding $\mathbf{f}_v$ and the sentence token  embedding $\mathbf{f}_x$, we first compute the precision (P) and recall (R) and then combine them to get the F1 score (F) as our fine-grained embedding matching score $\text{EMScore}(X, V)_{f}$ :

\vspace{-3mm}
\begin{equation}
	\label{equation: fine_P}
	\begin{footnotesize}
		P(X, V)_{f} = \frac{1}{|X|} \sum_{x_{i} \in X} \max _{v_{j} \in V} \mathbf{f}_{x_i}^{\top} \mathbf{f}_{v_j} ,
	\end{footnotesize}
	\vspace{-1.5mm}
\end{equation}

\vspace{-1mm}
\begin{equation}
	\label{equation: fine_R}
	\begin{footnotesize}
		R(X, V)_{f} = \frac{1}{|V|} \sum_{v_{j} \in V} \max _{x_{i} \in X} \mathbf{f}_{x_i}^{\top} \mathbf{f}_{v_j} ,
	\end{footnotesize}
	\vspace{-1mm}
\end{equation}

\vspace{-1.5mm}
\begin{equation}
	\label{equation: fine_F}
	\begin{footnotesize}
		\text{EMScore}(X, V)_{f} = 2 \frac{P \cdot R}{P+R} .
	\end{footnotesize}
	\vspace{-1mm}
\end{equation}
By such token-frame matching in the calculation of precision, it is easy to figure out which visual frame is aligned with a specific word. The precision evaluates the correctness of the caption, such as whether descriptions are related to the video content without incorrect details. Similarly, it is easy to figure out which word is aligned with a specific visual frame in the calculation of recall. The recall evaluates the comprehensiveness of the caption, such as whether the content in the video is described without omission. The F1 measure combines the evaluation of these two aspects.


\noindent\textbf{IDF Weighting}
A caption usually consists of two kinds of words: visual content words like nouns and function words like ``the'', ``and'', \etc. For these function words, it is hard to align them with the video frames and thus we should lower their importance weight during token-frame matching. Since the more visual-irrelevant words will appear more times in the whole caption corpus, \eg, the word ``a'' may appear in every sentence, we calculate the inverse document frequency (idf) to weigh the importance of each word and integrate it into EMScore.
Given a corpus $\left\{X^{(i)}\right\}_{i=1}^{N}$, the idf value of a token $x$ is:
\vspace{-3.5mm}
\begin{equation}
	\label{equ:compute idf}
	\begin{footnotesize}
		\operatorname{idf}(x)=-\log \frac{1}{N} \sum_{i=1}^{N} \mathbb{I}\left[x \in X^{(i)}\right] ,
	\end{footnotesize}
	\vspace{-3mm}
\end{equation}
where $\mathbb{I}[\cdot]$ is an indicator function. The special token [EOS] appears in each caption and Eq.~\eqref{equ:compute idf} will assign its weight as 0, while this token contains comprehensive contextual information of the whole sentence since it is used as the discriminative signal for classification during the pre-training in CLIP. To remedy this, we empirically set the idf value of the [EOS] token to the average value of the entire idf set.

After calculating the idf values, the Precision in Eq.~\eqref{equation: fine_P} is changed to:
\vspace{-4mm}
\begin{equation}
	\begin{footnotesize}
		\begin{aligned}
			P(X, V)_{f}= \frac{\sum_{x_{i} \in X} \operatorname{idf}\left(x_{i}\right) \max _{v_{j} \in V} \mathbf{f}_{x_i}^{\top} \mathbf{f}_{v_j}}{\sum_{x_{i} \in X} \operatorname{idf}\left(x_{i}\right)} .
		\end{aligned}
	\end{footnotesize}
	\vspace{-2mm}
\end{equation}
When calculating Precision and Recall, 
IDF is applied for $X$ and $V$, respectively. Note that idf weighting will not affect the calculation of Recall in Eq.~\eqref{equation: fine_R} since each frame is equally important. 

\vspace{-1mm}
\subsection{EMScore \& EMScore\_ref}
\vspace{-1mm}

\label{subsec:emscore}
When calculating EMScore, we do not need any reference and only use the video $V$. Specifically, EMScore is defined as the average of $\text{EMScore}_{c}$ and $\text{EMScore}_{f}$:
\vspace{-2.5mm}
\begin{equation}
	\begin{footnotesize}
		\label{equation: EMScore(X, V)}
		\begin{aligned}
			\text{EMScore}(X, V) = \frac{\text{EMScore}(X, V)_{c} + \text{EMScore}(X, V)_{f}}{ 2 } .
		\end{aligned}
	\end{footnotesize}
	\vspace{-1.5mm}
\end{equation}The score is in the range [-1, 1]. A higher EMScore indicates a better caption, as it is more consistent with the video.

When the reference caption $X^{*}$ is available, we can incorporate it to get EMScore\_ref. 
First, $\text{EMScore}(X, X^{*})$ are calculated as in Eq.~\eqref{equation: EMScore(X, V)} by replacing $V$ with $X^{*}$, and the ground truth embeddings are changed from the frame and video representations to word and text representations.
Second, we define enhanced EMScore\_ref as the average of $\text{EMScore}(X, V)$ and $\text{EMScore}(X, X^{*})$.

\vspace{-6mm}
\begin{equation}
	\begin{footnotesize}
		\begin{aligned}
			\text{EMScore\_ref}(X, V, X^{*}) = \frac{\text{EMScore}(X, V) + \text{EMScore}(X, X^{*}) }{ 2 } .
		\end{aligned}
	\end{footnotesize}
	\label{equation: EMScore(X, V, X*)}
\end{equation} If there are multiple reference sentences $\{X_{i}^{*}\}_{i=1}^{M}$,
$\operatorname{EMScore}\left(X, X^{*}\right)=\max \limits_{i} \operatorname{EMScore}\left(X, X_{i}^{*}\right)$.
In the following, unless otherwise specified, \text{EMScore} refers to $\text{EMScore(X, V)}$, and \text{EMScore\_ref} refers to $\text{EMScore(X, V, X*)}$.

\vspace{-1mm}
\section{The Collected Datasets}
\vspace{-1mm}
\subsection{The VATEX-EVAL Dataset}
\vspace{-1mm}

The VATEX-EVAL dataset is collected to evaluate the correlation of automatic metrics with human judgment.

\noindent\textbf{Candidate Caption Collection}
We use all 3000 validation videos from VATEX~\cite{DBLP:conf/iccv/WangWCLWW19} and collect a total of 18,000 candidate captions with 6 captions per video. 
To span the full range of caption quality, for each video, we collect three kinds of captions: one high-quality, two medium-quality, and three low-quality captions. Specifically, for high-quality captions (GT), they are randomly selected from original ground-truth reference captions; for medium-quality captions (Top-Down and ORG-TRL), they are generated from Top-Down~\cite{DBLP:conf/cvpr/00010BT0GZ18} and ORG-TRL~\cite{DBLP:conf/cvpr/ZhangSY0WHZ20} captioning models; for low-quality captions (AM\_1, AM\_2, AM\_3), they are selected from other videos in the VATEX validation dataset by adversarial matching. More details of the caption collection are in the Appendix.

\begin{figure}
	\centering
	\setlength{\abovecaptionskip}{1mm}
	\includegraphics[width=0.7\linewidth]{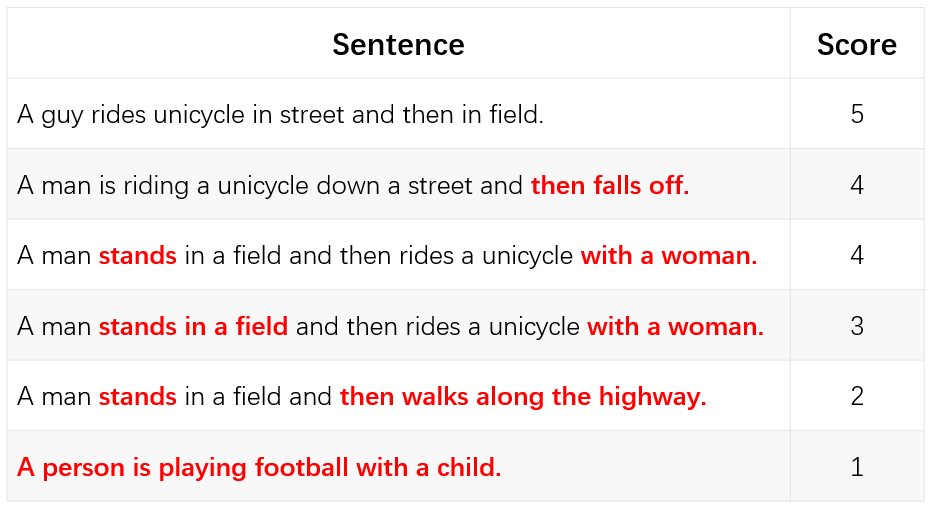}
	\caption{An annotation example for the VATEX-EVAL dataset. Incorrect details in the captions are red highlighted.}
	\label{fig:annotationexample}
	\vspace{-3mm}
\end{figure}

\begin{table}[]
	\centering
	\setlength{\abovecaptionskip}{1mm}
	\resizebox{0.45\textwidth}{!}{%
		\begin{tabular}{c|cccccc}
			\hline
			System        & GT    & Top-Down & ORG-TRL & AM\_1 & AM\_2 & AM\_3 \\ \hline
			Average Score & 4.750 & 3.920    & 4.003   & 3.916 & 3.854 & 3.793 \\ \hline
		\end{tabular}%
	}
	\caption{Average scores for the six different caption source.}
	\label{tab:human_average_scores}
	\vspace{-6mm}
\end{table}

\noindent\textbf{Human Evaluation Setup}
\label{subsection: Human Evaluation Setup}
To ensure high quality of the human evaluation, each candidate caption is scored by 3 English-speaking annotators, amounting to 54,000 human ratings. 
For each video, we ask 3 annotators to rate the consistency degrees between the captions between the video. The rate scales from 1 to 5 where 1 denotes inconsistent and 5 denotes consistent. Fig.\ref{fig:annotationexample} shows one example where incorrect details are red highlighted. Annotators are provided with detailed instruction (refer to Appendix), which is written to minimize subjectivity in annotations.


\noindent\textbf{Dataset Annalysis}
We demonstrate the reliability of our collected VATEX-EVAL dataset from two aspects. Firstly, to check the agreements among different annotators, we compute the Kendall and Spearman correlation coefficients, which are 0.568 and 0.628 respectively. 
These inter-annotator correlations indicate strong inter-annotator agreements.  
Secondly, Tab.\ref{tab:human_average_scores} presents the average annotation scores for the six candidate caption collection sources. 
The average score of the original ground-truth captions strongly outperforms those of all other caption types, which is in line with the fact that GT captions have the highest quality intuitively. 
The ORG-TRL model gets a higher annotation score than the Top-Down model, which is also positively correlated with model complexity.  
The three captions of adversarial matching also give reasonable and reliable scores in the order of the adversarial matching score. The above analysis proves that our annotations are reliable.



\begin{figure}
	\centering
	\setlength{\abovecaptionskip}{1mm}
	\includegraphics[width=0.7\linewidth]{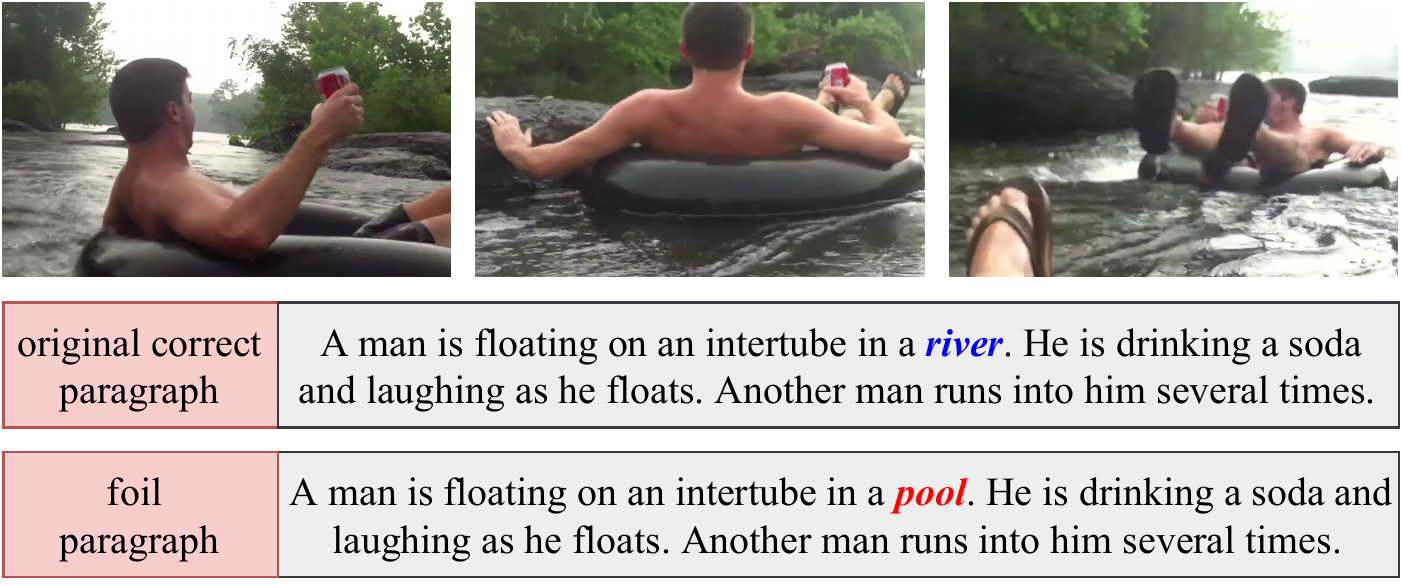}
	\caption{A correct-foil pair example for ActivityNet-FOIL.}
	\label{fig:ActivityNet-FOIL_example}
	\vspace{-6mm}
\end{figure}

\vspace{-1mm}
\subsection{The ActivityNet-FOIL Dataset}
\vspace{-2mm}

Prior work demonstrates that current captioning models generally generate “hallucinating” descriptions~\cite{rohrbach-etal-2018-object} that are not actually in the source visual scene. To test how sensitive EMScore is to identify foil captions that contain inaccurate visual concepts, we follow FOIL-COCO dataset~\cite{shekhar-etal-2017-foil} to change ActivityNet-Entities test dataset~\cite{8954099} for constructing an ActivityNet-FOIL dataset. In ActivityNet-Entities, each video has two corresponding paragraphs. We use one of two paragraphs to construct correct-foil pair, and use the other as a reference for reference-based metrics. Each paragraph has about 3 sentences in different time stamps, and a visual concept in each sentence is grounded to an annotation bounding box. A foil caption is created by replacing the original visual concept with a similar but false one.

Our data generation process has three main steps: 
First, we collect all visual concepts and filter out the ones with low frequency. Then we pair together words belonging to the same supercategory (such as river-pool, shirt-shoe, cat-dog). At last, we obtain 2,191 correct-foil pairs, in which each visual concept has approximately 13 foil ones.
Second, we replace a visual concept in the original correct caption with paired foil candidate to construct a candidate foil caption. Each correct caption has multiple candidate foil captions.
Third, for each correct caption, we mine the hardest foil caption by selecting the lowest perplexity candidate. 
Finally, we create 1900 correct-foil paragraph pairs, and at least one caption in the foil paragraph contains a foil visual concept.
As shown in Fig.\ref{fig:ActivityNet-FOIL_example}, it contains a correct-foil paragraph pair per video. We compute the accuracy of each metric in its capacity to assign a higher score to the correct candidate paragraph versus the foil. More details about the ActivityNet-FOIL collection can be seen in the Appendix. 

\vspace{-2.7mm}
\section{Experiments}
\vspace{-1mm}

We conduct experiments to evaluate our EMScore and EMScore\_ref on VATEX-EVAL (cf. Section~\ref{subsection: Results on VATEX-EVAL}) and ActivityNet-FOIL (cf. Section~\ref{subsection: Pairwise Ranking on ActivityNet-FOIL}) datasets. 
To measure caption-level human correlation, we compute Kendall’s correlation $\tau$ and Spearman’s rank correlation $\rho$.

We compare EMScore with four rule-based metrics, \eg, BLEU~\cite{DBLP:conf/acl/PapineniRWZ02}, ROUGE\_L~\cite{lin-2004-rouge}, METEOR~\cite{banerjee-lavie-2005-meteor} and CIDEr~\cite{7299087}\footnote{These metrics are implemented in MS COCO evaluation tool \url{https://github.com/tylin/coco-caption}.}, and two embedding-based metrics, \eg, BERTScore~\cite{DBLP:conf/iclr/ZhangKWWA20}\footnote{\url{https://github.com/Tiiiger/bert\_score}} and Improved BERTScore~\cite{DBLP:conf/acl/YiDH20}\footnote{\url{https://github.com/ck0123/improved-bertscore-for-image-captioning-evaluation}}. For these two embedding-based metrics, we utilize RoBERTa-base~\cite{Liu2019RoBERTaAR} as the backbone, and use F1-measure with idf and Recall with idf, respectively, which are the best setting in their paper.
For our EMScore and EMScore\_ref, they can also optionally combine with idf. Specifically, the training caption corpus from the source dataset (VATEX and ActivityNet) is used to calculate idf. For the value of $|V|$, we use all frames in video. The value of $h$ and $d$ are both 512.

\vspace{-2.5mm}
\subsection{Results on VATEX-EVAL}
\vspace{-1mm}

\label{subsection: Results on VATEX-EVAL}

\subsubsection{Ablation Study}
\vspace{-2mm}

\noindent\textbf{Effect of P, R, F, and idf weighting}
From Tab.\ref{tab:P_R_F_idf}, we can see that F1-measure has achieved relatively stable performance regardless of whether idf weighting is used or not.
After adding idf weighting, the performance on precision and F1-measure of EMScore is improved. The result proves that idf weighting is effective. 
The best performance is obtained under the combination of calculating F1-measure and using idf weighting. Therefore, in the following, we use F1-measure combined with idf weighting as the default setting.

\begin{table}[]
	\centering
	\setlength{\abovecaptionskip}{1mm}
	\resizebox{0.22\textwidth}{!}{%
		\begin{tabular}{lcc}
			\hline
			Metric                       & $\tau$          & $\rho$          \\ \hline
			\text{EMScore}$_{f}$ (P)     & 0.1843          & 0.2404          \\
			\text{EMScore}$_{f}$ (R)     & 0.2263          & 0.2946          \\
			\text{EMScore}$_{f}$ (F)     & 0.2228          & 0.2900          \\
			\text{EMScore}$_{f}$ (P-idf) & 0.2052          & 0.2674          \\
			\text{EMScore}$_{f}$ (R-idf) & 0.2263          & 0.2946          \\
			\text{EMScore}$_{f}$ (F-idf) & \textbf{0.2296} & \textbf{0.2989} \\ \hline
		\end{tabular}%
	}
	\caption{The performance difference of different calculation methods and the effect of idf weighting on fine-grained \text{EMScore}$_{f}$. $\tau$/$\rho$ indicates the Kendall/Spearman correlation, respectively. 
	}
	\label{tab:P_R_F_idf}
	\vspace{-3mm}
\end{table}

\begin{table}[]
	\centering
	\setlength{\abovecaptionskip}{1mm}
	\resizebox{0.28\textwidth}{!}{%
		\begin{tabular}{lllcc}
			\hline
			\# & Metric                            & GT   & $\tau$                              & $\rho$                              \\ \hline
			1  & \text{EMScore}$_{c}$              & V    & 0.2269                              & 0.2955                              \\
			2  & \text{EMScore}$_{f}$ (F-idf)      & V    & 0.2296                              & 0.2989                              \\
			3  & \text{EMScore} (F-idf)            & V    & \textbf{0.2324}                     & \textbf{0.3026}                     \\ \hline
			4  & \text{EMScore}$_{c}$              & X*   & \multicolumn{1}{l}{0.2390}          & \multicolumn{1}{l}{0.3104}          \\
			5  & \text{EMScore}$_{f}$ (F-idf)      & X*   & \multicolumn{1}{l}{0.2495}          & \multicolumn{1}{l}{0.3240}          \\
			6  & \text{EMScore} (F-idf)            & X*   & \multicolumn{1}{l}{\textbf{0.2550}} & \multicolumn{1}{l}{\textbf{0.3307}} \\ \hline
			7  & \text{EMScore\_ref}$_{c}$         & V+X* & 0.2738                              & 0.3548                              \\
			8  & \text{EMScore\_ref}$_{f}$ (F-idf) & V+X* & 0.2779                              & 0.3599                              \\
			9  & \text{EMScore\_ref} (F-idf)       & V+X* & \textbf{0.2863}                     & \textbf{0.3705}                     \\ \hline
		\end{tabular}%
	}
	\caption{The effect of different granularities embedding matching and the effect of different ground truths. GT, V, X* are denoted as ground truth, video, and reference. For X*, there is one reference. $\tau$/$\rho$ indicates the Kendall/Spearman correlation, respectively. 
	}
	\label{tab:Effect of Multi-grained}
	\vspace{-4mm}
\end{table}

\noindent\textbf{Effect of Different Granularities and Ground Truths}
\label{subsubsection: Effect of Different Granularity}
In Tab.\ref{tab:Effect of Multi-grained}, we first observe the impact of different granularities.
For EMScore which uses video as ground truth (GT), the fine-grained EMScore in line 2 achieves better results than the coarse-grained one in line 1. The result verifies our motivation that it is correct to consider the characteristics of the visual elements of the video over time in the video caption evaluation process. 
Moreover, the performance of the combination of two granularities in line 3 is further improved. The result verifies that multi-granularity combination is beneficial.
Next, we observe the impact of using different ground truths. When using both video and reference as GT, a better correlation result is achieved than using them alone. The result proves our conjecture that the information in the video and references are complementary, and additional use of references can bring information gain. Therefore we recommend using EMScore\_ref when references are available.

\vspace{-3.5mm}
\subsubsection{Comparsion with the other metrics}
\vspace{-1.5mm}

\label{subsubsection: Caption level Human correlation}

\begin{table}[]
	\centering
	\setlength{\abovecaptionskip}{1mm}
	\resizebox{0.47\textwidth}{!}{%
		\begin{tabular}{lcccccc}
			\hline
			\multirow{2}{*}{Metric}   & \multicolumn{2}{c}{No Ref} & \multicolumn{2}{c}{1 Ref} & \multicolumn{2}{c}{9 Refs} \\ \cline{2-7} 
			& $\tau$       & $\rho$      & $\tau$      & $\rho$      & $\tau$       & $\rho$      \\ \hline
			BLEU\_1                   & -            & -           & 0.1219      & 0.1591      & 0.289        & 0.3697      \\
			BLEU\_4                   & -            & -           & 0.0806      & 0.0881      & 0.216        & 0.256       \\
			ROUGE\_L                  & -            & -           & 0.1249      & 0.1631      & 0.2378       & 0.3085      \\
			METEOR                    & -            & -           & 0.1644      & 0.2149      & 0.2763       & 0.3574      \\
			CIDEr                     & -            & -           & 0.1732      & 0.2263      & 0.2781       & 0.3606      \\ \hline
			BERTScore (F-idf)         & -            & -           & 0.1824      & 0.2373      & 0.293        & 0.3775      \\
			Improved\_BERTScore (R-idf) & -            & -           & 0.1516      & 0.198       & 0.2442       & 0.3167      \\ \hline
			EMScore (F-idf)           & \textbf{0.2324}       & \textbf{0.3026}      & -           & -           & -            & -           \\
			EMScore\_ref (F-idf)      & -            & -           & \textbf{0.2863}      & \textbf{0.3705}      & \textbf{0.3681}       & \textbf{0.4719}      \\ \hline
		\end{tabular}%
	}
	\caption{Human correlation on the VATEX-EVAL dataset. $\tau$/$\rho$ indicates the Kendall/Spearman correlation respectively. 
	}
	\label{tab:correlation_res}
	\vspace{-4mm}
\end{table}

In the following experiment, we prove that our EMScore achieves higher human correlation and lower reference dependency, which benefits from the introduction of the video content. We also show that our metric is robust to quality drift and has a consistent system-level ranking with humans.

\begin{figure}
	\centering
	\setlength{\abovecaptionskip}{-0.5mm}
	\includegraphics[width=0.8\linewidth]{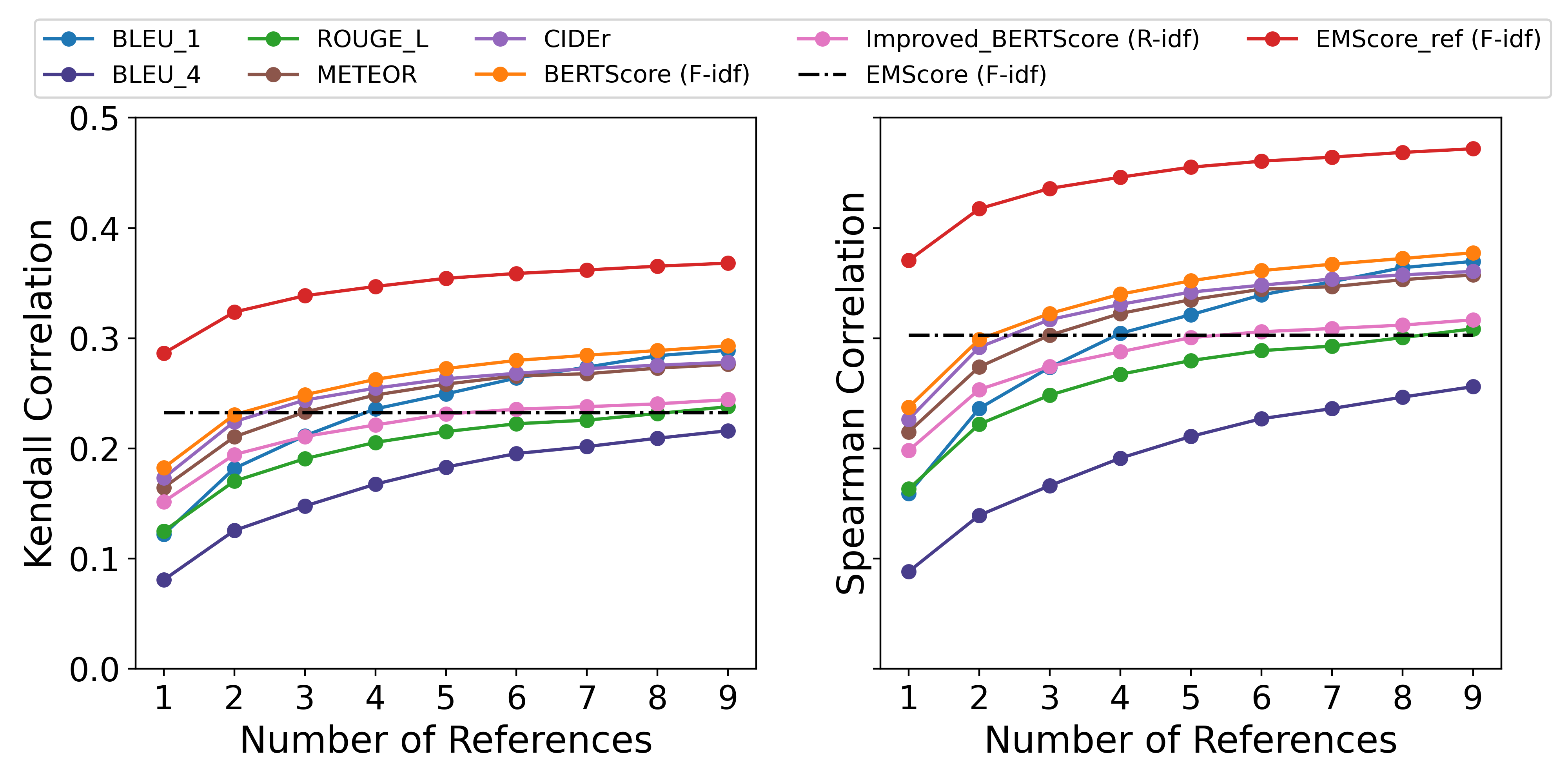}
	\caption{The Kendall and Spearman correlations between automatic metrics and human judgments with the different numbers of references on the VATEX-EVAL dataset. The dashed line indicates EMScore, which does not rely on any reference.}
	\label{fig:changing_reference_number}
	\vspace{-5mm}
\end{figure}

\noindent\textbf{High Human Correlation} 
Tab.\ref{tab:correlation_res} shows the correlation results of metrics with 0, 1, 9 references per candidate caption.
We have several observations as follows:
(1) In situations where no references are available, our EMScore still works well, and achieves surprisingly competitive results. The result demonstrates the advantage of taking video content into account, while other reference-based metrics cannot handle this situation; 
(2) When using the same number of references (e.g., 1 or 9), our EMScore\_ref outperforms other prior metrics by a large margin. The comparison results prove that our metric achieves higher human correlation and we propose a more reliable metric.

\noindent\textbf{Low Reference Dependency} 
The Kendall and Spearman correlations between automatic metrics and human judgments with the different numbers of references are shown in Fig.\ref{fig:changing_reference_number}.
Our EMScore without any references can achieve competitive results, compared with reference-based metrics which need at least 4 or 5 references, such as BLEU\_1 and Improved\_BERTScore.  
Besides, our EMScore\_ref with only one reference can achieve comparable results with reference-based metrics, which need at least 8 or 9 references, such as CIDEr and BERTScore. 
The results show that our metric has lower reference dependency, which benefits from the introduction of video content in evaluating.

\begin{table*}[]
	\centering
	\setlength{\abovecaptionskip}{1mm}
	\resizebox{0.7\textwidth}{!}{%
		\begin{tabular}{lrllrllrllrllrl}
			\hline
			System   & \multicolumn{2}{c}{Human} &  & \multicolumn{2}{c}{EMScore (F-idf)} &  & \multicolumn{2}{c}{EMScore\_ref (F-idf)} &  & \multicolumn{2}{c}{CIDEr}    &  & \multicolumn{2}{c}{BERTScore (F-idf)} \\ \hline
			GT       & 0.937        & (1)        &  & 0.581             & (1)             &  & 0.639                & (1)               &  & 0.178 & \textcolor{red}{(2)} &  & 0.498      & \textcolor{red}{(3)}     \\
			ORG-TRL  & 0.751        & (2)        &  & 0.539             & (2)             &  & 0.606                & (2)               &  & 0.185 & \textcolor{red}{(1)} &  & 0.527      & \textcolor{red}{(1)}     \\
			Top-Down & 0.730        & (3)        &  & 0.530             & (3)             &  & 0.591                & (3)               &  & 0.173 & \textcolor{red}{(3)} &  & 0.515      & \textcolor{red}{(2)}     \\
			AM\_1    & 0.729        & (4)        &  & 0.522             & (4)             &  & 0.584                & (4)               &  & 0.146 & (4)                  &  & 0.464      & (4)                      \\
			AM\_2    & 0.714        & (5)        &  & 0.515             & (5)             &  & 0.571                & (5)               &  & 0.140 & (5)                  &  & 0.451      & (5)                      \\
			AM\_3    & 0.698        & (6)        &  & 0.512             & (6)             &  & 0.566                & (6)               &  & 0.134 & (6)                  &  & 0.447      & (6)                      \\ \hline
		\end{tabular}%
	}
	\caption{System-level ranking on the VATEX-EVAL dataset. Nine references are used in the reference-based metrics and our EMScore\_ref. Each column for the metrics in the table gives the score for each system and the ranking of the six systems. The red fonts highlight denote that the metric's ranking is inconsistent with humans.}
	\label{tab:system-level-table}
	\vspace{-4mm}
\end{table*}

\noindent\textbf{Robust to Quality Drift}
\label{subsubsection: Robustness to Quality Drift}
It is extremely important for metrics to deal with quality drift, since the quality of generated captions can vary significantly across different video captioning models. 
To assess the robustness of metrics to quality drift, we create biased sets from our annotated VATEX-EVAL dataset by sampling candidate captions of different quality levels with different probabilities. Specifically, the annotation score of each caption ranges from 1 to 5. We then create 5 biased sets, indexed by the variable $I \in\{1,2, \cdots, 5\}$. For the $I^{th}$ set, we sample the candidate captions whose annotation score is $k$ with a probability of $\frac{1}{|I-k|+1}$, where $k \in\{1,2, \cdots, 5\}$.

\begin{figure}
	\centering
	\setlength{\abovecaptionskip}{1mm}
	\includegraphics[width=0.95\linewidth]{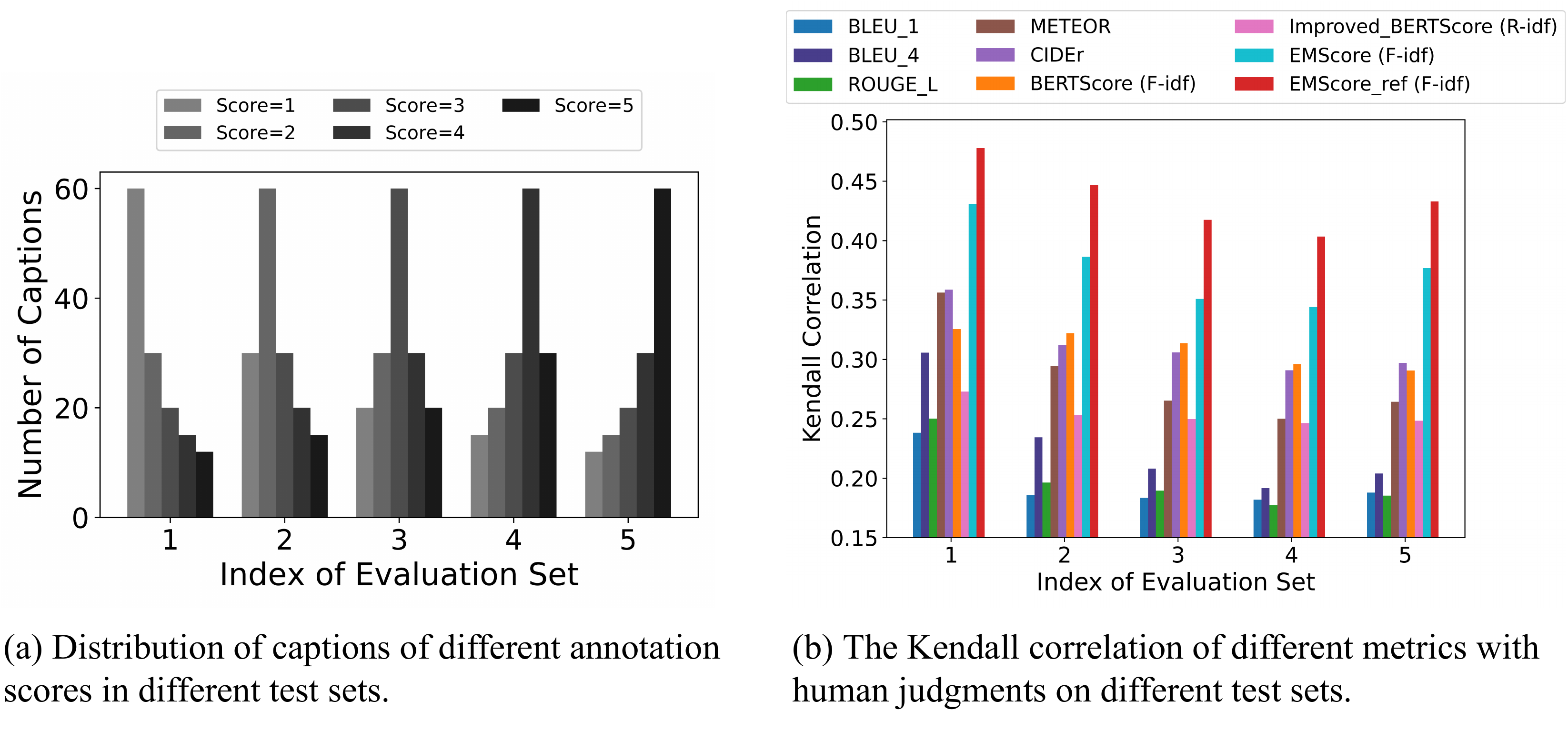}
	\caption{Robustness of metrics over different caption quality biased sets. 
		One reference is used in reference-based metrics and our \text{EMScore\_ref}.
	}
	\label{fig:Quality_Drift}
	\vspace{-5mm}
\end{figure}

In this way, the 5 sets have different distributions of candidate captions with different qualities, as shown in Fig.\ref{fig:Quality_Drift} (a). 
We compute the Kendall correlation between different metrics and human judgments on the 5 sets. 
One reference is used in the reference-based metrics and our EMScore\_ref.
Fig.\ref{fig:Quality_Drift} (b) shows that: (1) Our metrics EMScore and \text{EMScore\_ref} have a higher correlation than other metrics on all biased sets, which proves that our metrics are robust to the quality drift; (2) We find that rule-based metrics, e.g., BLEU\_4, perform much better on low-quality captions (set 1) than on high-quality (set 5). With the development of video captioning, they will become increasingly unreliable because they struggle to judge high-quality captions.

\begin{figure*}
	\centering
	\setlength{\abovecaptionskip}{0mm}
	\includegraphics[width=0.8\linewidth]{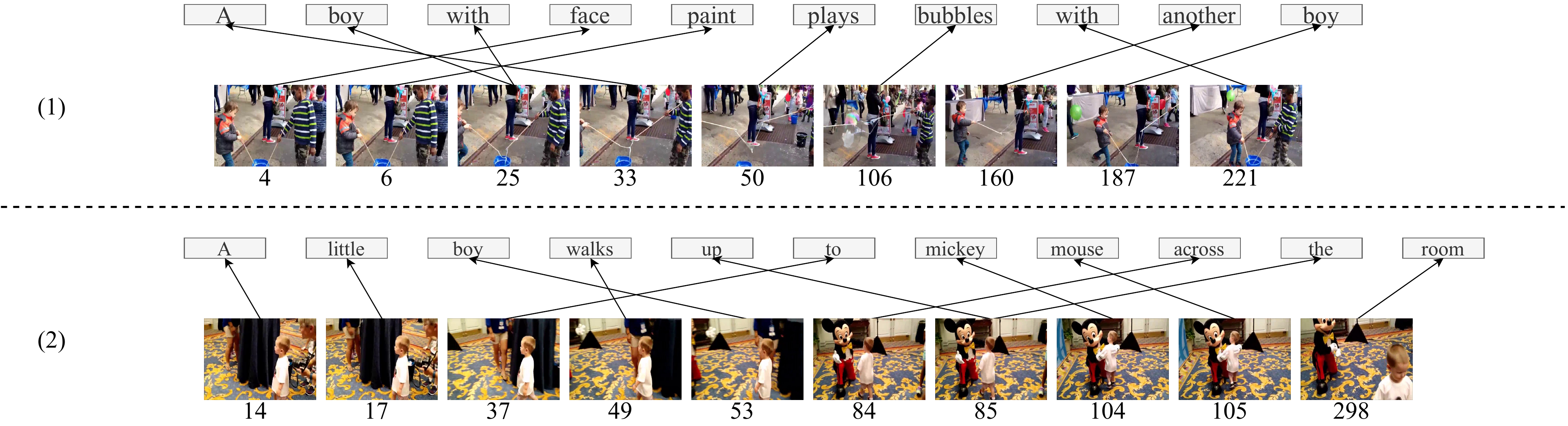}
	\caption{EMScore precision visualization. Each token is matched to the most similar frame. The temporal index is shown under the frame.} 
	\label{fig:fine_grained_matching}
	\vspace{-6mm}
\end{figure*}

\noindent\textbf{System-level Ranking on VATEX-EVAL}
\label{subsubsection: System-level Ranking on VATEX-EVAL}
Video captioning researchers generally report system-level scores to verify the effectiveness of their methods, so, it is essential to measure the metric's system-level human correlation. A reliable metric is expected to have the same system ranking as humans. In Tab.\ref{tab:system-level-table}, we compare the ranking of six system average scores rated by metrics and humans on the VATEX-EVAL datasets. All the metric scores are scaled to [0, 1], including human scores. For the human scores, the GT system gets the highest score, and following by ORG-TRL, Top-Down, AM\_1, AM\_2, AM\_3.  We use red fonts to highlight that the metric's ranking is inconsistent with human ranking. 
We can see that CIDEr and BERTScore cannot correctly rank GT, ORG-TRL, and Top-Down systems, \eg, they give the highest score to the ORG-TRL system instead of GT. Our EMScore and EMScore\_ref are consistent with human ranking. The result shows that our EMScore and EMScore\_ref are more reliable in system-level evaluation than other metrics, and it will be beneficial for the video captioning development.

\begin{table}[]
	\centering
	\setlength{\abovecaptionskip}{1.0mm}
	\resizebox{0.42\textwidth}{!}{%
		\begin{tabular}{lc|lc}
			\hline
			Metric            & Accuracy(\%) & Metric                            & Accuracy(\%) \\ \hline
			BLEU\_1           & 60.11        & $\text{EMScore}_{c}$              & 87.95        \\
			BLEU\_4           & 66.11        & $\text{EMScore}_{f}$ (F-idf)      & 90.32        \\
			ROUGE\_L          & 56.74        & EMScore (F-idf)                   & 89.47        \\
			METEOR            & 72.89        & $\text{EMScore\_ref}_{c}$         & 90.21        \\
			CIDEr             & 77.89        & $\text{EMScore\_ref}_{f}$ (F-idf) & 93.00        \\ \cline{1-2}
			BERTScore (F-idf) & 86.68        & EMScore\_ref (F-idf)              & 92.42        \\ \hline
		\end{tabular}%
	}
	\caption{Pairwise ranking accuracy on ActivityNet-FOIL dataset.}
	\label{tab:ActivityNet-table}
	\vspace{-6mm}
\end{table}

\vspace{-5mm}
\subsubsection{EMScore Visualization}
\vspace{-2mm}

\label{subsubsection: EMScore Visualization}
Fig.\ref{fig:fine_grained_matching} visualizes how fine-grained EMScore matches the most similar visual elements to the tokens (as the calculation of precision).
For the first example, ``bubbles'' is occurred in the 106th frame, ``another boy'' is occurred in the 160th and 187th frames, and compared with other frames, ``face paint'' appears in a larger proportion in the 4th and 6th frames. 
For the second example, the visual concept ``boy'' appears as the main visual element in the 53th frame, so the token 'boy' matches this frame instead of 84th$\sim$298th frames where multiple visual elements appear. 
Compared with coarse-grained embedding matching, our fine-grained one can take into account the characteristic of the video, and provide more interpretability for EMScore.

\vspace{-2mm}
\subsection{Experiments on ActivityNet-FOIL}
\vspace{-2mm}

\label{subsection: Pairwise Ranking on ActivityNet-FOIL} 
To test the capability of EMScore to identify ``hallucinating'' captions, 
we compute the accuracy of pairwise ranking for each evaluation metric in their capacity to assign a higher score to the correct candidate paragraph versus the foil on the ActivityNet-FOIL dataset.
Each candidate paragraph has multiple captions, so we first compute the caption score, then calculate the overall score of the paragraph as the average score of multiple captions. The ground truth for each caption is obtained by cutting the video and the reference paragraph into multiple segments and reference captions, respectively, according to the timestamp of the candidate captions. 
The accuracy results are shown in the Tab.\ref{tab:ActivityNet-table}, we have the following findings:
(1) Even without any reference, our \text{EMScore} outperforms all reference-based metrics. Moreover, our EMScore reaches a noteworthy improvement in terms of accuracy by about 3\% compared to the best prior metric (BERTScore 86.68\%). The excellent result proves that it is effective to take the video content as ground truth in the hallucination caption identification;
(2) When enhanced by the reference, our EMScore\_ref$_f$ achieves the highest accuracy (93.00\%); 
(3) Due to the large changes in the visual scene of the video in the ActivityNet dataset, it will be more effective to consider fine-grained embedding matching than coarse-grained one. At the same time, the multi-granularity combination does not bring performance improvement, and the results suggest that it is sufficient to use fine-grained matching alone for videos with large visual scene changes.

\vspace{-3mm}
\section{Conclusion}
\vspace{-1.3mm}

In this paper, we have conducted a systematic study on the video captioning evaluation metrics. First, to solve the drawbacks of reference-based metrics, we have proposed a novel video captioning evaluation metric \text{EMScore} by measuring the consistency between the video and caption. 
Second, we have collected two datasets (VATEX-EVAL and ActivityNet-FOIL) to systematically analyze the reliability of the existing metrics. The VATEX-EVAL experiments have demonstrated that our EMScore has a higher human correlation and lower reference dependency. Moreover, it is robust to quality drift, and consistent with humans on the system-level ranking. The ActivityNet-FOIL experiments have shown that our EMScore is sensitive to identifying ``hallucinating'' captions. 

\noindent\textbf{Limitations.}  
EMScore is an embedding-based metric, and relies on the performance of used vision-language pre-trained (VLP) model.
More reliable evaluation scores can be obtained by leveraging better VLP models to extract better representations.
More discussion about the effect of VLP models is in the Appendix.


\noindent\textbf{Acknowledgements.} This work is supported by Beijing Natural Science Foundation (Grant No. JQ21017), the national key R\&D program of china (No. 2020AAA0105702), the Natural Science Foundation of China (Grant No. 61972397, 62036011, 62192782, 61721004, U19B2038, 61906192), the Key Research Program of Frontier Sciences, CAS, Grant No. QYZDJ-SSW-JSC040, the University Synergy Innovation Program of Anhui Province under Grants GXXT-2019-025, the Science and Technology Service Network Initiative, CAS (Grant No. KFJ-STS-SCYD-317).

{\small
	\bibliographystyle{ieee_fullname}
	\bibliography{egbib,shiyaya_reference}
}

\end{document}